%% file: PaperForReview.tex
\documentclass[10pt,twocolumn,letterpaper]{article}

\usepackage{wacv}              

\usepackage{graphicx}
\usepackage{algorithm}
\usepackage{algpseudocode}
\usepackage{amsmath}
\usepackage{multirow}
\usepackage{amssymb}
\usepackage{booktabs}
\usepackage{comment}

\usepackage[pagebackref,breaklinks,colorlinks]{hyperref}

\usepackage[capitalize]{cleveref}
\crefname{section}{Sec.}{Secs.}
\Crefname{section}{Section}{Sections}
\Crefname{table}{Table}{Tables}
\crefname{table}{Tab.}{Tabs.}

\def\wacvPaperID{1537} 
\def\confName{WACV}
\def\confYear{2025}

\begin{document}

\title{To Ask or Not to Ask? \\ Detecting Absence of Information in Vision and Language Navigation}
\author{ Savitha Sam Abraham$^{\ast}$, Sourav Garg$^{\ast}$,  Feras Dayoub$^{\ast}$ \\
$^{\ast}$ Australian Institute for Machine Learning \\
The University of Adelaide\\
{\tt\small \{savitha.samabraham, sourav.garg, feras.dayoub\}@adelaide.edu.au}
}

\maketitle

\begin{abstract}
Recent research in Vision Language Navigation (VLN) has overlooked the development of agents' inquisitive abilities, which allow them to ask clarifying questions when instructions are incomplete. This paper addresses how agents can recognize ``when" they lack sufficient information, without focusing on ``what" is missing, particularly in VLN tasks with vague instructions. Equipping agents with this ability enhances efficiency by reducing potential digressions and seeking timely assistance. The challenge in identifying such uncertain points is balancing between being overly cautious (high recall) and overly confident (high precision). We propose an attention-based instruction-vagueness estimation module that learns associations between instructions and the agent's trajectory. By leveraging instruction-to-path alignment information during training, the module's vagueness estimation performance improves by around 52\% in terms of precision-recall balance. In our ablative experiments, we also demonstrate the effectiveness of incorporating this additional instruction-to-path attention network alongside the cross-modal attention networks within the navigator module. Our results show that the attention scores from the instruction-to-path attention network serve as better indicators for estimating vagueness.

\end{abstract}

\begin{figure}
    \centering
    \includegraphics[scale=0.28]{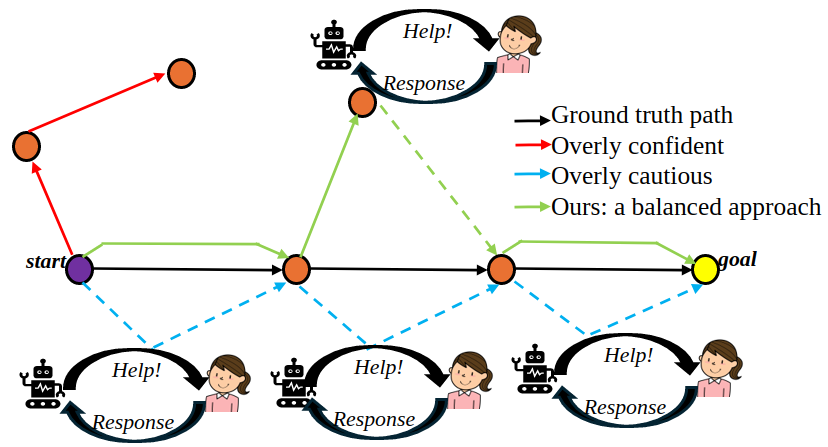}
    \caption{Navigation agent paths using different instruction-vagueness estimation approaches: (1) Overly confident approach (red) that rarely seeks help, (2) Overly cautious approach (blue) that seeks help very often and (3) Our balanced approach (green) that seeks timely assistance. Dashed arrows indicate movements made with external assistance (best viewed in color).}
    \label{fig:enter-label}
\end{figure}
\section{Introduction} \label{sec:intro}
Vision Language Navigation (VLN) is the task of enabling a robot to navigate an environment based on a given instruction. There has been substantial research in VLN in the recent past \cite{wu2024vision}. The majority of previous research has focused on training large neural models for VLN task using synthetic datasets like R2R \cite{anderson2018vision} that contained step-by-step instructions and Reverie \cite{qi2020reverie} that had high-level instructions, among others. All these models operate under the assumption that agents are designed to act independently without human intervention. Even when dealing with abstract or high-level instructions, agents are trained to explore and make decisions they deem best for progressing towards their goals. There is no provision for indicating uncertainty or indecision about the next course of action. Typically, designers impose an upper time limit in these approaches, prompting agents to cease operations after a predefined number of moves, regardless of goal achievement.

When a robot transitions to a real-world environment, it often encounters instructions from various individuals, each with unique communication styles. Adapting to this diversity poses a challenge in generalizing across different instructing styles. While some individuals may prefer concise instructions, others might delve into details but present them in a disorganized manner. A second person unfamiliar with the instructor may find it challenging to follow the instructions easily. In such situations, humans typically seek clarification by asking questions like, ``\textit{I'm now in front of the kitchen. Where should I go next?}". Implementing a similar ability is a desired feature in intelligent robots to tackle uncertainty due to vagueness in the input instruction. When a robotic agent is posed with an unclear instruction, it may get stuck at some point, or it may keep exploring, taking very long paths before reaching the goal, or it may never reach the goal. Rather than waiting for the robot to encounter failure, it is preferable if the robot can be proactive and seek help when it is approaching failure.

We are interested in the task of Under-specified Vision Language Navigation (ULN) as outlined in \cite{feng2022uln}. ULN introduced a nuanced modification to the VLN task by incorporating a new dimension of information completeness. In this paper, we focus on the task of instruction-vagueness estimation within the context of ULN. Instruction-vagueness (IV) estimation assesses the uncertainty arising from vague input instructions in the decision-making process of a VLN agent. In essence, an IV estimation module addresses the VLN agent's query:  ``\textit{Do I possess enough information to proceed with my next move or do I seek assistance?}". A key requirement for this module is to find the right balance (see Figure \ref{fig:enter-label}) between being overly cautious and frequently seeking help, which guarantees task completion (high recall), and being overly confident, potentially resulting in excessive exploration without finishing the task (high precision). The contributions of this paper are as follows:

\begin{itemize}
    
   \item We introduce an attention-based \textbf{instruction-vagueness (IV) estimation} module and integrate it into an existing VLN model. The module takes the instruction, the path followed thus far, and the proposed next move as its inputs and decides at each time step \textit{whether to follow the VLN model's suggestion or to request assistance}.

  \item We introduce a \textbf{pre-training task} that helps \textit{identify important parts of the instructions needed for predicting actions}. By incorporating this instruction to path alignment information into the Instruction-Vagueness (IV) module, we significantly improve its ability to detect points of uncertainty, enhancing precision-recall balance.

\end{itemize}

\section{Related Work}\label{sec:rel}

In this section we provide a brief survey of recent works related to vision language navigation and uncertainty arising from vagueness in input instruction. 

\subsection{Vision Language Navigation}
Existing methods in VLN vary depending on the information used to predict the next location to move to. Chen et al. \cite{chen2021history} introduced HAMT (History Aware Multimodal Transformer), a framework that retains a record of all previously visited and observed locations as historical data. This information, along with the input instruction is then utilized to determine the next destination, typically chosen from among the neighboring locations of the current position. VLN-DUET (Dual Scale Graph Transformer), as described in \cite{chen2022think}, employs a detailed representation of the current observation derived from objects in the image instead of a coarse representation of the current location. Additionally, it also maintains a topological map of the nodes visited thus far as in \cite{chen2021history} to determine the subsequent node for navigation. This next node may include any unvisited nodes observed thus far, expanding beyond the immediate local neighbors (performing global action planning). Recent studies aim to create large-scale VLN datasets \cite{rawal2024aigen, chen2022learning} to enhance the generalizability of VLN agents. Speaker follower models \cite{fried2018speaker} and instruction following and generation models \cite{wang2023lana} focus on refining instruction generation and interpretation abilities of the agent. Numerous researchers have also made significant efforts to enhance vision-language alignment in VLN by introducing a variety of pre-training tasks \cite{du2024delan}, \cite{deguchi2024language}, \cite{zhang2024navhint}. With the emergence of Large Language Models (LLMs), recent advancements in VLN have increasingly incorporated LLMs in navigation \cite{zhou2024navgpt, qiao2023march}. 

In this paper, we evaluate the effectiveness and generalizability of our proposed IV estimation module by incorporating it into two navigators, VLN-DUET and HAMT. We first examine how well these navigators, trained with detailed instructions from the \textit{R2R} dataset \cite{anderson2018vision}, adapt to vague instructions. We integrate our IV estimation module into both to \textit{recognize potential absence of critical information}, prompting them to seek assistance.

\subsection{Knowing when you don't know}
 ULN \cite{feng2022uln} as mentioned in Section \ref{sec:intro}, introduced the dataset based on R2R for the problem of under-specified VLN. ULN modified the R2R dataset by adding a high-level instruction, that intentionally omitted certain details, alongside the original fine-grained instruction for each trajectory. These high-level, vague instructions corresponding to the fine-grained instructions were manually created and validated. The paper handled vagueness in instruction by introducing two separate modules - an instruction classifier and an uncertainty estimation module. The instruction classifier classified an instruction as either high or low level. The uncertainty estimation module predicts a score indicating the uncertainty of the next move, using candidate moves weighted by likelihood and attention scores. In this paper, we introduce an attention-based IV estimation module that assesses the uncertainty arising from instruction vagueness in the predictions of a VLN model. We demonstrate that the alignment between the given instruction and the path taken so far is an effective metric for gauging IV. In contrast to the approach in \cite{feng2022uln}, where an instruction classifier labels an instruction as low or high-level (indicative of potential vagueness) based solely on the instruction content, we contend that \emph{vagueness arises not merely from the instruction itself, but from the interaction between the instruction and the surrounding environment}. 

Vision Dialog Navigation (VDN) involves enabling robots to seek help during navigation \cite{thomason2020vision}. A related work \cite{zhu2021self} introduced an MLP-based uncertainty estimator that identified uncertainty points based on the navigator's current state and was trained using pseudo-labels of uncertainty derived from the entropy of the navigator's predicted distribution. We compare this with our IV model, which uses instruction-path attention, and show that our approach more accurately indicates uncertainty and generalizes across different navigators. We demonstrate that relying on the predicted distribution's entropy can be misleading, as the navigator might confidently make incorrect predictions about the next move.

Another recent work that explored uncertainty estimation is \textit{KnowNo} \cite{ren2023robots}.  It utilised conformal prediction to gauge uncertainty in the next-step prediction of a LLM-based planner. Uncertainty due to instruction ambiguity is studied in \textit{CLARA} \cite{park2023clara}, where a LLM-based framework was used to discern whether a provided user input instruction is ambiguous or infeasible. It adopts the method of self-consistency check involving sampling multiple responses from the LLM for the same query, and assessing the consistency among these responses. This approach is similar to the conventional ensemble method for uncertainty estimation \cite{lakshminarayanan2017simple}. It is not efficient to implement an ensemble-based approach for VLN considering the complexity of the task, which typically comprises individual networks for text encoding, image encoding, topological graph encoding and cross-attention. It requires training all these layers or a selected sub-network (text encoder) as done in \cite{feng2022uln} separately, freezing all other layers. Our aim here is to avoid re-training any part of the VLN model, and limit the training to just the IV module.

\section{Method}\label{sec:method}
\subsection{Problem Background and Formulation} \label{sec:pf}
In a VLN setup for discrete environments, the environment at any time step \( t \) is represented as an undirected graph \( G_t = (V_t, E_t) \). \( G_t \) represents the map of the environment explored so far with \( V_t = \{N_i\}_{i=1}^K \), being the set of nodes or locations observed so far and \( E \) denoting the connectivity between the locations. Nodes are flagged as either visited or navigable (potential next nodes). The graph starts with the agent's initial position and updates at each time step to include the current position and new navigable nodes. 

The input instruction  \( I = \{w_i\}_{i=1}^L \) is a sequence of word embeddings, with \( L \) words. The agent’s goal is to navigate from its current position to the goal location described in the instruction. The VLN model includes a text encoder \( f_{\text{text}} \) (typically a multi-layer transformer) that generates contextual representation of the instruction, \( \hat{I} \). The agent also has a cross-attention network,  \( f_{\text{cross-attention}} \), that learns a graph-aware and instruction-aware representation for each node in the graph, represented as \( \hat{G_t} = (\hat{V_t}, E_t) \). It also learns the attention weights of visual inputs on language, represented as \( \alpha_t \). The action predictor network, \(f_{\text{actionPredictor}} \) predicts the subsequent node the agent will move to at each time step by assigning a likelihood score to each navigable node in \(\hat{G_t}\), represented as \( \beta_t \). 
\begin{align}
\hat{I} &= f_{\text{text}}(I) \\
\hat{G_t}, \alpha_t &= f_{\text{cross-attention}}(\hat{I}, G_t) \\
\beta_t &= f_{\text{actionPredictor}}(\hat{I}, \hat{G_t})
\end{align}
\(f_{\text{actionPredictor}} \) thus predicts the next node to move to at each time step, eventually resulting in a sequence of nodes or the path, \( P = \{N_i\}_{i=1}^{L_{p}} \), traversed by the agent to execute the task. \(L_p\) is the length of the path generated. The segment of the path traversed up to time \(t\) is denoted by \(P_t\). 

An Instruction-Vagueness (IV) estimation module,  \(f_{\text{IV}} \), computes for each time step, \( t \), the uncertainty, \( score_{\text{t}}^{\text{uncertainty}} \), arising from vagueness in the instruction, \( \hat{I} \), in the prediction of \( \beta_t \) by \(f_{\text{actionPredictor}} \). 
\begin{figure*}
    \centering
    \includegraphics[scale=0.3]{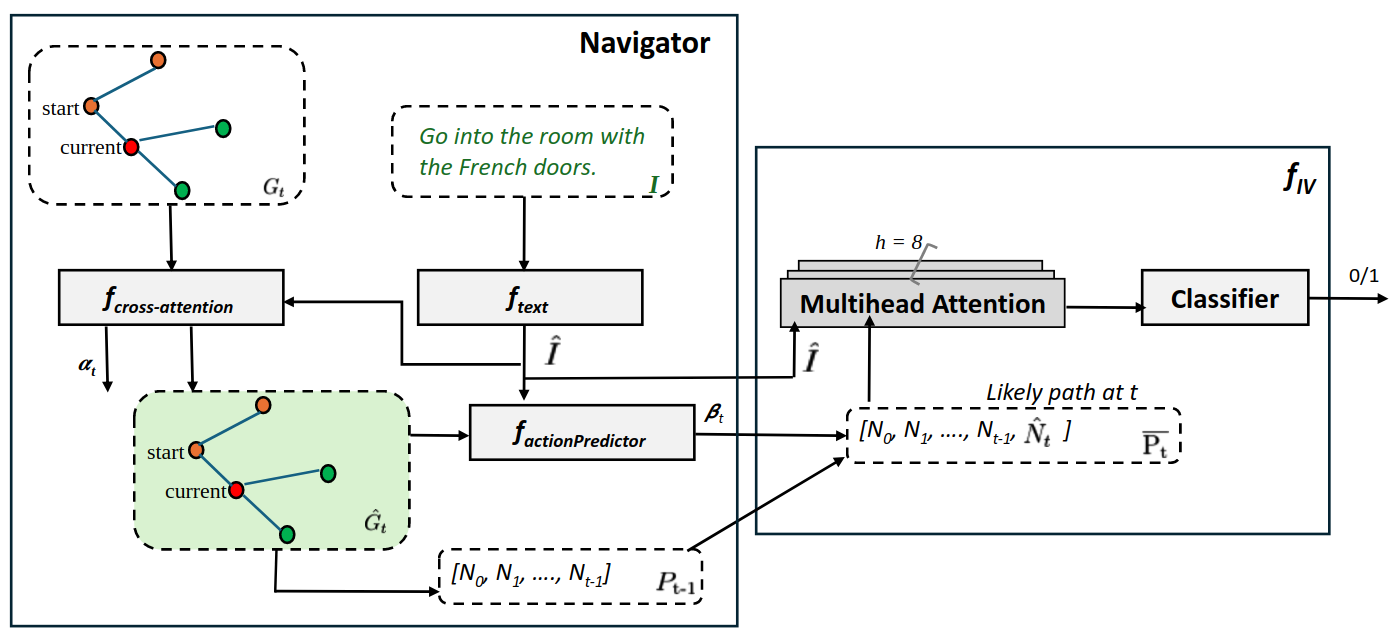}
    \caption{Interaction between the navigator and the IV module: The IV module receives the encoded instruction and the path taken so far with a suggestion for the next move from the navigator. It predicts the certainty in the navigator's next move suggestion.}
    \label{fig1}
\end{figure*}
\subsection{Our approach: Instruction Vagueness Estimation} \label{sec:iv}
The architecture of \(f_{\text{IV}} \) is shown in Figure \ref{fig1}. The vagueness in instruction at any point during traversal is modeled as a function of both the instruction and the path being undertaken. If node \(\hat{N}_{t}\) is predicted as the most likely node to move to at time step \(t\) by \(f_{\text{actionPredictor}}\), 
\begin{align}
\overline{\mathrm{P_{t}}} = P_{\text{t-1}} &+ \hat{N}_{t} \\
score_{\text{t}}^{\text{uncertainty}} &= f_{\text{IV}} (\overline{\mathrm{P_{t}}},  \hat{I}) 
\end{align}
Here, \(P_{\text{t-1}}\) denotes the path taken by the agent until time \((t-1)\). Appending \(\hat{N}_{t}\), the next move suggested by \(f_{\text{actionPredictor}} \) to it gives \(\overline{\mathrm{P_{t}}}\), the most likely path up to time \(t\).  The goal of \(f_{\text{IV}} \) is to determine the uncertainty arising from instruction vagueness in the prediction of \(f_{\text{actionPredictor}}\). 
\(f_{\text{IV}} \) employs a multi-head attention (MHA) network \cite{vaswani2017attention}  to learn the alignment between this anticipated path at time \(t\) and the instruction.
\begin{equation}
    \text{Instr\_Path\_Attn}_t = \text{MultiHead}(\hat{I}, \overline{\mathrm{P_{t}}}, \overline{\mathrm{P_{t}}})
\end{equation}
\begin{equation}
\text{MultiHead}(\hat{I}, \overline{\mathrm{P_t}}, \overline{\mathrm{P_t}}) = \text{Concat}(\text{head}_1, \ldots, \text{head}_h)W^O 
\end{equation}
\begin{equation}
\text{head}_i = \text{Attention}(\hat{I}W_i^Q, \overline{\mathrm{P_t}}W_i^K, \overline{\mathrm{P_t}}W_i^V)
\end{equation}
\begin{equation}
\text{Attention}(\hat{I}, \overline{\mathrm{P_t}}, \overline{\mathrm{P_t}}) = \text{softmax}\left(\frac{\hat{I}(\overline{\mathrm{P_t}})^T}{\sqrt{d_k}}\right)\overline{\mathrm{P_t}}
\end{equation}
Here,  \(Instr\_Path\_Attn_t\) is an enhanced representation that captures the alignment between \( \hat{I} \) and the likely path, \(\overline{\mathrm{P_t}}\).  \( W_i^Q, W_i^K, W_i^V \) are learned projection matrices for each attention head \( i \), \( W^O \) is a learned output projection matrix, \( h \) denotes the number of attention heads ($=8$), and \( d_k \) represents the dimensionality of the instruction embedding ($=768$). The enhanced representation, \(Instr\_Path\_Attn_t\), is then passed onto a binary classifier, a linear transformation layer that produces two scores for the two classes ``certain" and ``uncertain". 
\begin{equation}
    \text{score}_{\text{t}}^{\text{uncertainty}} = W \cdot \text{Instr\_Path\_Attn}_t + b
\end{equation}
where $W$ is a $2 \times d_k$ weight matrix, and $b$ is the bias. If $score_{\text{t}}^{\text{uncertainty}}$ indicates ``uncertainty", the agent is prompted to seek assistance; otherwise, it proceeds with the suggested move. In this study, we assume the existence of an oracle in place of a human who provides assistance. When the agent requests assistance, this oracle provides the optimal move — the one that guides the agent along the shortest path to its goal from its current position.

\noindent \textbf{Supervision}: 
Training \(f_{IV}\) network requires ground truth labels indicating whether the prediction made by \(f_{\text{actionPredictor}}\) is uncertain ($= 1$) or not ($=$ 0). Since, we do not have these labels, we approximate it in two ways.  
\begin{itemize} \label{sec:gp}
    \item \textbf{Based on ground truth path ($label_{\text{GP}}$)} : 
    If the move predicted by the VLN agent at time step $t$ is same as the ground truth move at $t$, it is considered ``certain", else, uncertain.  This way of approximating uncertainty labelling is commonly observed in literature \cite{feng2022uln}.
    \item \textbf{Based on instruction-path alignment ($label_{\text{IP}}$)}: In this paper, we explore using instruction-to-path alignment for uncertainty labeling. This approach assumes the existence of information about instruction-to-path alignment, which associates each node in the path with a specific sub-instruction.  Given a ground truth path, $GP = \{N_0, N_1, \dots\}$ and each instruction represented as a list of sub-instructions,  \( SI = \{S_1, S_2, \dots, S_m\}\), the instruction to path alignment maps $GP$ to $SI$. This mapping is denoted as \(Rel\_SI(N_t)\), which returns the sub-instruction from $SI$ that is aligned with $N_t$ in $GP$.  This mapping is utilized to label uncertainty arising from instruction vagueness as follows:
    \begin{equation}
    \text{uncertainty}_t = 
    \begin{cases}
        0 & \text{if } \text{Rel\_SI}(N_t) \text{ in input instruction} \\
        1 & \text{otherwise}
    \end{cases}
\end{equation}
In essence, this labeling checks at each time step whether the relevant sub-instruction is present in the input instruction.  If not, it indicates that the input instruction is vague and lacks necessary information relevant for making the prediction at that point.
\end{itemize}

\noindent \textbf{Loss function}: The loss function used in either case is the binary cross entropy loss (BCE) \cite{rumelhart1986learning}. 

\subsection{Pre-training task: Relevant Instruction Span Identification}

\begin{figure}
    \centering
    \includegraphics[scale=0.28]{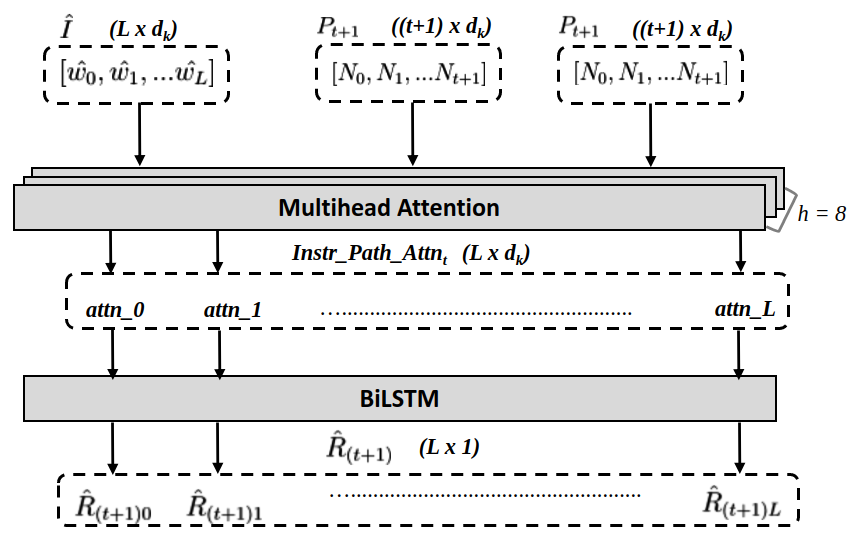}
    \caption{Pre-training: Architecture of the network that learns to identify the most relevant span or chunk in $\hat{I}$ that influenced the last move made, $N_{\text{t+1}}$.}
    \label{fig:2}
\end{figure}
We propose a pre-training task that serves to initialize the weights of the multihead attention network in $f_{\text{IV}}$. The task is a instruction to path alignment task. The objective of this task is to predict the segment/chunk of the instruction that is most relevant when predicting the agent's move at time-step \(t\), given the path taken so far, \(P_t\). It is formulated as a sequence-to-sequence classification problem, where each token in the input instruction is classified as 0 or 1 based on its relevance. The task is defined as:
\begin{align}
    \hat{R_t} &= f_{\text{pretrainAlign}} (P_t,  \hat{I})  
\end{align}
In this context, $\hat{I}$, a $L \times d_k$ dimensional representation of the instruction, is transformed into a $L \times 1$ dimensional relevance sequence. $L$ denotes the number of tokens in the instruction. The output sequence indicates the relevance of each token (assigned 1 if relevant, otherwise 0) in predicting the move at time step $t$.

The architecture of the network for the pretraining task is illustrated in Figure \ref{fig:2}. It consists of a multihead attention layer followed by a BiLSTM network  \cite{huang2015bidirectional} that maps the refined representation that incorporates instruction to path alignment information, to the relevance sequence \cite{sutskever2014sequence}. We opt for a BiLSTM rather than a more complex transformer model due to data limitations.

\noindent \textbf{Supervision}: We leverage the instruction to path alignment information in the dataset, Fine-Grained R2R (FGR2R) \cite{hong2020sub}, for training \(f_{\text{pretrainAlign}}\). Each fine-grained instruction \(I_{\text{orig}}\) in FGR2R is segmented into sub-instructions or chunks and each node from the ground truth path is mapped to the most relevant sub-instruction within the instruction. Such a  mapping, denoted as \(Rel\_SI(N_t)\), provides information about the sub-instruction from \(I_{\text{orig}}\) that is aligned with any node $N_t$ in the ground truth path. 
The network,  \(f_{\text{pretrainAlign}}\), is trained independently, keeping the weights of the different components in the navigator - $f_{\text{text}}$, $f_{\text{cross-attention}}$ and others frozen. It may be noted that the navigator acts in a teacher-forcing manner - that is, the move it makes at any time $t$ is always the ideal move or the move at time step $t$ in the ground truth path. This is because, the instruction-path alignment information in FGR2R is with respect to the ground truth paths in the dataset.   

\noindent \textbf{Loss function}: The network is trained using a combination of BCE loss and Dice loss \cite{milletari2016v}. While BCE loss ensures accurate probability prediction for each token independently, Dice loss encourages accurate identification of boundaries of relevant span within the instruction.
\section{Experiments} \label{sec:exp}
\subsection{Research Questions}
The following are the research questions we aim to address through our experiments:
\begin{enumerate}
    \item \textbf{RQ1}: How does the performance (precision-recall balance) of our proposed instruction-to-path attention-based instruction vagueness estimator compare to existing approaches? 
    \item \textbf{RQ2}:  How does utilizing different pseudo-labels for uncertainty - $label_{\text{IP}}$,  $label_{\text{GP}}$ and entropy-based label affect the performance of IV module?
    \item \textbf{RQ3}: Does incorporating instruction to path alignment knowledge through pre-training of multi-head attention layer improve the performance of IV module? 
\end{enumerate}

\subsection{Dataset} \label{sec:data}
The original R2R dataset \cite{anderson2018vision} is divided into four principal splits: \texttt{train}, \texttt{val\_seen}, \texttt{val\_unseen}, and \texttt{test}. The \texttt{val\_seen} split contains environments similar to those in the \texttt{train} split, whereas \texttt{val\_unseen} and \texttt{test} feature trajectories in entirely new, unobserved environments. The ULN dataset \cite{feng2022uln}, features a manually created coarse-grained or high-level instruction,  \(I_{\text{short}}\), along with the original fine-grained instruction,  \(I_{\text{orig}}\) for each trajectory in \texttt{val\_seen} and \texttt{val\_unseen}.
We also utilize the instruction to path alignment information for each trajectory from the FGR2R variant of R2R \cite{hong2020sub}. Next, we describe the data used for training and evaluation in our experiments.

\noindent \textbf{\(f_{\text{pretrainAlign}}\) train/val data}: \(f_{\text{pretrainAlign}}\) training exclusively uses the \texttt{train} split of R2R, which comprises approximately 11,400 trajectories. We divide this training set into training and validation subsets in an 80 to 20 ratio and do not use trajectories from \texttt{val\_seen} and \texttt{val\_unseen} during pre-training. 

\noindent \textbf{\(f_{\text{IV}}\) train/val data}: We train \( f_{\text{IV}} \) using approximately 2,000 trajectories from the \texttt{val\_seen} split, providing \(I_{\text{orig}}\) as input instruction for 50\% of the trajectories and \(I_{\text{short}}\) for the remainder. This approach ensures a balanced dataset enabling \( f_{\text{IV}} \) to discern between the two types of instructions during training. 

\noindent \textbf{\(f_{\text{IV}}\) test data}: We evaluate our instruction vagueness estimator on trajectories in \texttt{val\_unseen}, where \(I_{\text{orig}}\) is used as the input instruction for 50\% of the trajectories and \(I_{\text{short}}\) for the remaining trajectories.
We do not use the \texttt{test} split of R2R as it lacks annotation for \(I_{\text{short}}\) in it.

It is important to acknowledge that the data set available for training \( f_{\text{IV}} \) is limited as it is costly to manually annotate a given path with different styles of instructions. 


\subsection{Implementation Details}
The model \(f_{\text{pretrainAlign}}\) is trained for 7000 iterations with a batch size of 8, learning rate of 1e-4. Equal weights are given to both dice and BCE loss. The model \(f_{\text{IV}}\) is trained for 1000 iterations with a batch size of 8, learning rate of \mbox{1e-4}. The model is trained on imitation learning. Both models are trained with AdamW optimizer \cite{loshchilov2017decoupled} on a single GPU - NVIDIA GeForce RTX 4090.

\subsection{Baseline} \label{sec:sup}
The uncertainty estimation module in \cite{feng2022uln} is defined as follows:
\begin{align}
score_{\text{t}}^{\text{uncertainty}} &= f_{\text{Base}} (\alpha_t, \beta_t)
\end{align}
\(\alpha_t\) and \(\beta_t\) are as defined in Equations 2 and 3 in Section \ref{sec:pf}. This approach leverages \(\alpha_t\), the visual attention on instruction from the VLN model, and \(\beta_t\), which are the likelihood scores that the VLN agent assigns to potential next moves at each time step. These two components are concatenated and fed into a classifier, which then assesses the uncertainty in the predictions made by the VLN model. The approach uses supervised learning with pseudo-labels for uncertainty derived from ground truth path - $label_{\text{GP}}$. We denote this method as \(f_{\text{Base}}\). Comparing to \(f_{\text{Base}}\) enables us to assess whether the attention scores from the navigator serve as a better indicator of vagueness than the instruction-to-path attention scores learned by \(f_{\text{IV}}\). 

The next baseline is the uncertainty estimator used in VDN task \cite{zhu2021self}, defined as:
\begin{align}
score_{\text{t}}^{\text{uncertainty}} &= f_{\text{VDN}} (\alpha_t)
\end{align}
Here, pseudo-labels for uncertainty are derived from the entropy of the navigator's predicted distribution, $\beta_t$ (if entropy is close to that of a uniform distribution, i.e., within a predefined threshold, \(\epsilon\) $\in [0,1]$, it is considered uncertain). Comparing with 
\(f_{\text{VDN}}\) also allows us to evaluate the effectiveness of entropy as an uncertainty indicator versus using pseudo-labels based on the ground truth move $label_{\text{GP}}$ and alignment information $label_{\text{IP}}$.

The next baseline involves conformal prediction, as outlined in \cite{ren2023robots}. This method utilizes a calibration set to determine a threshold, \(\theta\), for the probability assigned to the most probable next move by the VLN agent. The threshold is set as a function of user-defined error tolerance (is set as 0.9 in our experiments). Moves are accepted if their probabilities fall within this threshold; otherwise, they are deemed uncertain. We denote this method as \(f_{\text{CP}}\). 

\subsection{Evaluation Metrics} \label{sec:eval}
We evaluate the Instruction Vagueness (IV) module using a precision-recall balance metric:
\begin{equation}
    \text{Balance} =  \frac{\text{Precision} - \text{Recall}}{\text{Precision} + \text{Recall}}
\end{equation}
Balance $\in [-1,1]$; negative values indicate higher recall, positive values indicate higher precision, and 0 indicates a balance with Precision $=$ Recall. The ground truth labels for uncertainty is based on $label_{\text{GP}}$.

\begin{table}[h]
\centering
\begin{tabular}{c|ccc}
\cline{1-4}
\textbf{Model} & \textbf{Instruction Style} & \textbf{SPL (\%)} & \textbf{NE (m)}   \\ 
\cline{1-4}
\multirow{2}{*}{DUET} & \(I_{\text{orig}}\) & 54.46 & 3.86   \\ 
 & \(I_{\text{short}}\) & 47.35 & 5.30 \\ 
\cline{1-4}
\multirow{2}{*}{HAMT} & \(I_{\text{orig}}\) & 43.82 & 5.46   \\ 
 & \(I_{\text{short}}\) & 31.58 & 7.11 \\ 
\bottomrule
\end{tabular}
\caption{Generalization to coarse-grained or high-level instruction for DUET and HAMT.}
\label{tab:1}
\end{table}

Predicting `uncertain' prompts the agent to seek help, opting for guidance from the oracle rather than following its original predicted move. This intervention can help redirect an agent that might be straying from the correct path back towards the intended route. We examine how assistance from the oracle enhances performance metrics such as Success Path Length (SPL) and Navigation Error (NE), which are standard metrics in VLN evaluations \cite{anderson2018evaluation}. NE measures the distance between the agent’s final position and the target location in meters, while SPL evaluates both the success of reaching the target and the path efficiency.

\input{floats/table_hamt}
\subsection{Results}\label{sec:res}
Our aim is not to outperform existing VLN models in SPL and NE but to address the issue of identifying information gaps during navigation. The improvement in SPL and NE are achieved with oracle assistance and is used as an indicator of timely assistance. We base our experiments on two VLN models, though we believe the IV module can be integrated into any VLN model with components as shown in Figure \ref{fig1}.

\noindent \textbf{Generalization to coarse-grained or high-level instruction}:
\noindent Table \ref{tab:1} shows the performance of VLN models, DUET \cite{chen2022think} and HAMT \cite{chen2021history}, trained on R2R dataset on \textbf{\(f_{\text{IV}}\) test data}. The first row shows the SPL and NE when \(I_{\text{orig}}\) is used as the input instruction for all the cases. The second row shows the results when \(I_{\text{short}}\) is used as the input instruction. We see a drop in performance, confirming that the models trained on a specific instruction style (\(I_{\text{orig}}\) style) struggle to generalize to a different instruction style (\(I_{\text{short}}\) style). It may also be noted that DUET performs better than HAMT in either cases. 

As mentioned in Section \ref{sec:data}, \textbf{$f_{\text{IV}}$ test data} comprises a balanced dataset with respect to two instruction styles: 50\% of instances have high-level instructions as input, denoted as $\textit{Inst}_{\text{short}}$, and the remaining have fine-grained instructions as input, denoted as $\textit{Inst}_{\text{orig}}$. It is important to note that while $I_{\text{orig}}$ is fine-grained and contains more details than $I_{\text{short}}$, it may also benefit from additional details, as indicated by potential improvements that can be made in both SPL and NE with $I_{\text{orig}}$ instructions (see Table \ref{tab:1}). These results suggest that the generated paths differ from the ground truth paths.

The goal of IV module is to identify uncertain moves and reduce the drop that is seen in SPL and NE with the change in instruction style. For instance, with IV support, DUET is expected to achieve an SPL above 47.35 and NE below 5.30 when \(I_{\text{short}}\) is used as the input instruction. If effective, DUET with IV may reach an SPL higher than 54.46, surpassing the SPL with original fine-grained instructions, showing that timely assistance can guide the agent closer to the ground truth path.
 
Table \ref{tab:2} compares various variants of our proposed approach \textbf{$f_{\text{IV}}$} with baseline approaches, when integrated with two navigators - DUET and HAMT. The column $\textit{Orig/Short}$ shows the percentage of instances in each category (\(Inst_{\text{Orig}}\) versus \(Inst_{\text{Short}}\)) where an oracle intervention was recommended by the instruction vagueness estimation module.



\begin{figure*}
    \centering
    \includegraphics[scale=0.3]{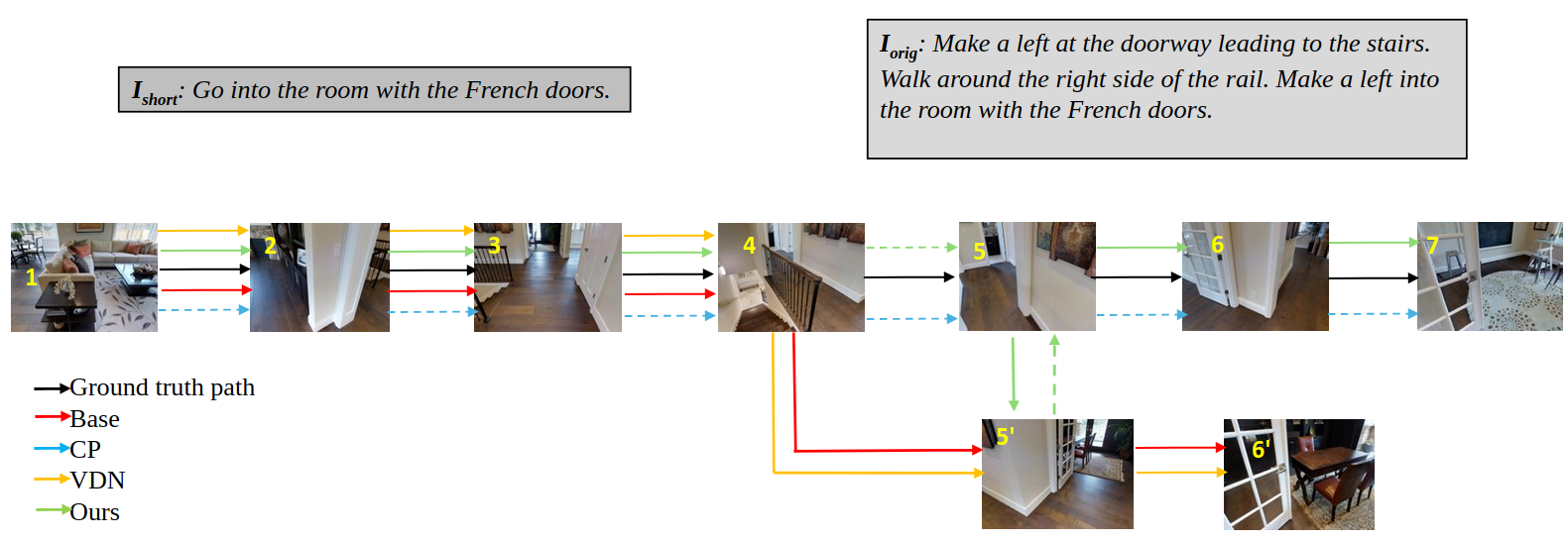}
    \caption{A navigation example with the input $I_{short}$ showing the agent trajectories when supported by $f_{CP}$ (blue), $f_{Base} (red)$ and our approach  $f_{IV(GP+pretrain)}$ (green). Dashed arrows indicate movements with oracle intervention (best viewed in color).}
    \label{fig:qual}
\end{figure*}

\begin{figure}
    \centering
    \includegraphics[scale=0.325]{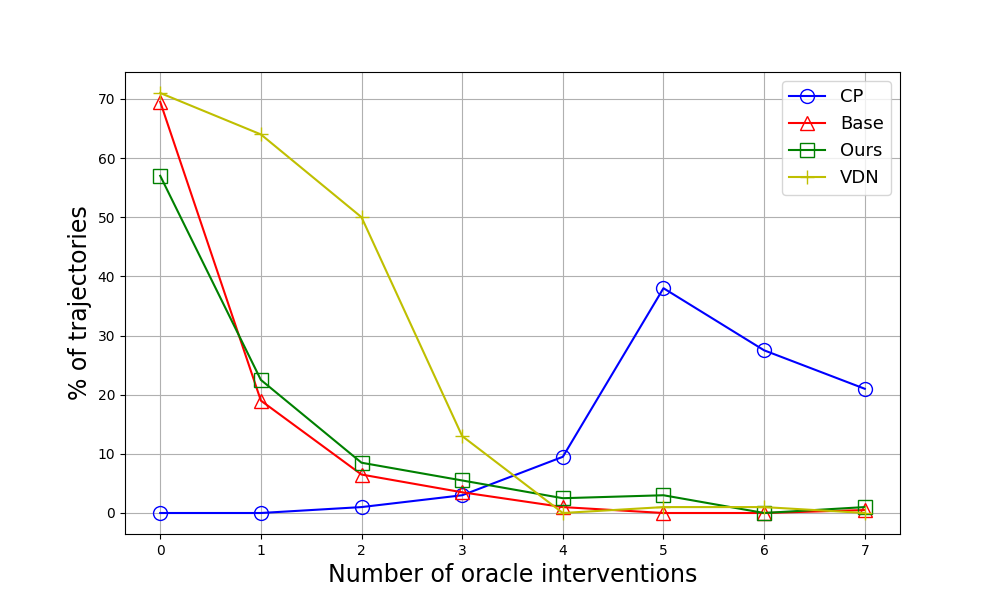}
    \caption{Number of oracle interventions vs. trajectories for \(f_{\text{CP}}\), \(f_{\text{Base}}\), \(f_{\text{VDN}}\) and ours - \(f_{\text{IV(GP+pretrain)}}\) (best viewed in color).}
    \label{fig:plot}
\end{figure}

\noindent \textbf{Comparing to baselines (RQ1)}: Table \ref{tab:2} illustrates that $f_{\text{CP}}$, an approach based on conformal prediction for uncertainty estimation, exhibits very high recall. It tends to be overly cautious, prompting intervention almost at every time-step, regardless of the input instruction style. This resulted in a high SPL and low NE. 

Next, we compare $f_{\text{Base}}$ with $f_{\text{IV(GP)}}$, a variant of $f_{\text{IV}}$. Both these models are trained with uncertainty labels based on ground truth paths ($label_{\text{GP}}$). It can be seen that both models perform better with HAMT than with DUET. This can be attributed to the fact that HAMT makes more incorrect predictions and hence $f_{\text{Base}}$ and $f_{\text{IV(GP)}}$ when integrated with HAMT, observes more cases of uncertainty during its training. The baseline $f_{\text{Base}}$, relying on attended representations $\alpha_t$ from the navigation model, shows more imbalance between precision and recall - with higher precision compared to recall, indicating excessive confidence in the navigation agent's decisions at each time-step. The results indicate that incorporating an explicit attention network in $f_{\text{IV}}$ to learn instruction-to-path attention representations improves precision-recall balance in uncertainty estimation compared to using the attended representation $\alpha_t$ from the navigation agent. 

\noindent \textbf{Comparing different pseudo-labels for uncertainty (RQ2)}: We examine the impact of using uncertainty labels derived from instruction to path alignment information ($label_{\text{IP}}$) instead of those derived from ground truth paths ($label_{\text{GP}}$). To do this, we compare the models \(f_{\text{IV(GP)}}\) and \(f_{\text{IV(IP)}}\). A notable finding is the distinct treatment of instances from the two styles, with more instances of oracle intervention suggested in $\textit{Inst}_{\text{short}}$ than in $\textit{Inst}_{\text{orig}}$. This results from $label_{\text{IP}}$, which is not agnostic to the instruction-style. It influences the IV estimation module to associate ``uncertain" label more with instructions of style \(I_{\text{short}}\) than with \(I_{\text{orig}}\). However, \(I_{\text{orig}}\) may also benefit from additional details as shown in our experiments, refer Table \ref{tab:1}. The significant increase in SPL and reduction in NE indicate its effectiveness in identifying uncertain points in $\textit{Inst}_{\text{short}}$ trajectories more efficiently, striking a balance between precision and recall. This demonstrates that such instruction-to-path alignment information can be a better indicator to identify missing information in instructions at any given time point. DUET showed best results with the variant \(f_{\text{IV(IP)}}\).

We compare $f_{\text{VDN}}$ using entropy-based pseudo-labels with $f_{\text{IV(GP)}}$ and $f_{\text{IV(IP)}}$. The results show that entropy-based labeling doesn't generalize well across navigators, especially with HAMT, where navigation errors are made with high confidence, causing pseudo-labels to indicate `certainty' even in uncertain cases.


\noindent \textbf{Impact of pre-training MHA in IV (RQ3)}: It is not feasible to assume that $label_{\text{IP}}$ is available for new data used to train \(f_{\text{IV}}\). Instead, we leverage instruction-to-path alignment information available from existing data (the FGR2R dataset) to pre-train the Multihead Attention (MHA) network in \(f_{\text{IV}}\). Table \ref{tab:2} demonstrates that \(f_{\text{IV(GP+pretrain)}}\), where $f_{IV}$ is trained with $label_{GP}$ and the MHA was initialized with pre-trained weights, exhibits improvement in SPL, NE and precision-recall balance compared to \(f_{\text{IV(GP)}}\), which uses an MHA initialized with random weights. These findings indicate that having an MHA network trained on the instruction-to-path alignment task can help enhance its performance in the estimation of instruction vagueness. 

\noindent \textbf{Number of oracle interventions}: Figure \ref{fig:plot} maps the percentage of trajectories to the number of times help was requested within a trajectory when the different models are integrated with DUET. Both $f_{Base}$ and our method reduces the number of questions per trajectory, with our approach requesting more assists than $f_{Base}$, as expected. Although $f_{VDN}$ has more interventions  per trajectory, the lower SPL and NE (see Table \ref{tab:2}) achieved despite more number of questions indicate that our approach is more effective in seeking timely assistance.

\noindent \textbf{Qualitative Results (an example with DUET)}: Figure \ref{fig:qual} illustrates an example with the instruction $I_{short}$: “\textit{Go into the room with the French doors}”. At time step 4, there is a point of uncertainty because the agent observes two rooms with French doors. At this point, both $f_{Base}$ and $f_{VDN}$ confidently moves to the nearest room with French doors but ends up at the wrong location (red, yellow paths). Our approach (green path) reaches the target after two interventions. In contrast, $f_{CP}$ is overly cautious, seeking assistance at every step (blue path).

\section{Conclusion and Future Scope}
\label{sec:conc}

In this paper, we presented a method to estimate uncertainty from vague instructions in under-specified vision-language navigation tasks. The approach focuses on aligning instructions with the generated path to indicate instruction vagueness.
Looking ahead, we aim to further develop this alignment-aware approach to precisely identify ``what information is exactly missing". 

\section{Acknowledgement}
This work was supported by the Centre for Augmented Reasoning, an initiative by the Department of Education, Australian Government.

{\small
\bibliographystyle{ieee_fullname}
\bibliography{PaperForReview}
}

\end{document}

%% file: floats/table_hamt.tex
\begin{table*}[h]
\centering
\begin{tabular}{c|cccccccc}
\cline{1-8}
\textbf{Method} & \textbf{Model} & \textbf{SPL (\%)} & \textbf{NE (m)} & \textbf{Precision (\%)} & \textbf{Recall (\%)} & \textbf{Balance} & \textbf{Orig/Short (\%)}  \\ 
\cline{1-8}
\multirow{6}{*}{DUET} 
& \(f_{\text{CP}}\) & 96.40 & 0.24 & 36.4791 & 99.0147 & -0.4615 & 100/100  \\ 
& \(f_{\text{Base}}\) & 52.10 & 4.04 & 72 & 11.1455 & 0.7319 & 28/33 \\ 
& \(f_{\text{VDN}}\) & 55.86 & 3.58 & 47.1962 & 15.2798 & 0.5108 & 64/65 \\ 
& \(f_{\text{IV(GP)}}\) & 57.86 & 3.51 & 62.5 & 14.5631 & 0.6220 & 30/42  \\ 
& \(f_{\text{IV(GP+pretrain)}}\) & 58.10 & 3.31 & 63.0681 & 17.5078 & 0.5654 & 40/46  \\ 
& \textbf{\(f_{\text{IV(IP)}}\)} & \textbf{75.24} & \textbf{2.16} & 37.5680 & 39.0566 & \textbf{-0.0194} & 26/79  \\ 
\cline{1-8} 
\multirow{6}{*}{HAMT} 
& \(f_{\text{CP}}\) & 88.34 & 1.29 & 54.4993 & 88.4773 & -0.2376 & 100/100 \\ 
& \(f_{\text{Base}}\) & 67.29 & 3.34 & 64.8230 & 49.6610 & 0.1324 & 81/ 92\\ 
& \(f_{\text{VDN}}\) & 38.38 & 5.63 & 41.5841 & 8.1237 & 0.6731 & 54/47\\ 
& \(f_{\text{IV(GP)}}\) & 68.82 & 3.21 & 55.4414 & 48.6486 & 0.0652 & 89/89 \\ 
& \textbf{\(f_{\text{IV(GP+pretrain)}}\)} & \textbf{75.84} & \textbf{2.32} & 51.2520 & 58.1439 & \textbf{-0.0629} & 90/96 \\
& \(f_{\text{IV(IP)}}\) & 70.32 & 3.35 & 39.2733 & 45.0396 & -0.0683 & 27/79 \\ 
\bottomrule
\end{tabular}
\caption{Results comparing baselines \(f_{\text{CP}}\), \(f_{\text{Base}}\), \(f_{\text{VDN}}\) with different variants of the proposed IV module (\(f_{\text{IV(GP)}}\) and \(f_{\text{IV(IP)}}\) : IV trained with pseudo-labels $label_{\text{GP}}$ and $label_{\text{IP}}$ respectively, \(f_{\text{IV(GP+pretrain)}}\): IV with MHA initialized to pretrained weights and finetuned with pseudo-label $label_{\text{GP}}$) on VLN-DUET and VLN-HAMT. We highlight the best variant: highest SPL, lowest NE and best balance (value closest to 0), with more emphasis on balance.}
\label{tab:2}
\end{table*}